%% file: ltc-task.tex
\documentclass[10pt, a4paper]{article}
\usepackage[utf8]{inputenc}
\usepackage[T1]{fontenc}
\usepackage{ltc05}
\usepackage{fancyvrb}
\usepackage{url}

\usepackage{graphicx}
\usepackage{listings}
\usepackage{tabularx}
\title{Open Challenge for Correcting Errors of Speech Recognition Systems} 

\name{Marek Kubis$^{\ast}$, Zygmunt Vetulani$^{\ast}$, Mikołaj Wypych$^{\dagger}$, Tomasz Ziętkiewicz$^{\dagger,\ast}$, \\ \bf \large} 

\address{ $^{\ast}$Adam Mickiewicz University \\
               Umultowska 87, 61-614 Pozna\'n, Poland \\ 
               \{mkubis,vetulani\}@amu.edu.pl \\ \\
               $^{\dagger}$Samsung Poland R\&D Institute \\ 
               Pl. Europejski 1, Warszawa, Poland \\
               \{m.wypych,t.zietkiewic\}@samsung.com}

\abstract{The paper announces the new long-term challenge for improving the performance 
of automatic speech recognition systems.
The goal of the challenge is to investigate methods of correcting the recognition results
on the basis of previously made errors by the speech processing system.
The dataset prepared for the task is described and evaluation criteria are presented.
}

\begin{document}

\maketitleabstract

\section{Introduction}

The rise in popularity of voice-based virtual assistants such as Apple's Siri, Amazon's Alexa, Google
Assistant and Samsung Bixby imposes high expectations on the precision of automatic speech
recognition (ASR) systems. 
Scheduling a meeting at the incorrect time, sending a message to a wrong person or
misinterpreting the command for the home automation system can cause severe losses to the
user of a virtual assistant.
The problem is even more apparent in case of deep-understanding systems
supposed to work in very difficult audibility condition, and where ASR errors can appear fatal 
for the end users. This is the case of systems for crisis situation management
(e.g. \newcite{Vetulani10b}) where low quality
and emotional voice input can generate a real challenge for speech recognition systems.
Furthermore, successful 
integration of ASR solutions with the very demanding AI systems will depend on the degree
of being able to take into consideration the non-verbal elements of utterances (prosody). 
Hence, despite significant improvements to the speech recognition
technology in recent years it is now even more important to search for new methods of
decreasing the risk of being misunderstood by the system.

One of the methods that can be used to improve the performance of a speech recognition system is to 
force the system to learn from its own errors. This approach transforms the speech recognition system
into a self-evolving, auto-adapting agent.
The objective of this challenge is to investigate to what extent this technique can be used to improve the
recognition rate of the speech processing system.

In order to make the challenge approachable by participants from outside the speech recognition
community and to encourage contestants to use a broad range of machine learning and natural
language engineering
methods that are not specific to the processing of spoken language, we provide the dataset
that consists solely of:

\begin{enumerate}
  \item \emph{Hypotheses} -- textual outputs of the automatic speech recognition system.
  \item \emph{References} -- transcriptions of sentences being read to the automatic speech recognition system.
\end{enumerate}
Thus, the goal of the contestants is to develop a method that improves the result of
speech recognition process on the basis of the (erroneous)
output of the ASR system and the correct human-made transcription without access to the speech
recordings.

\begin{figure}[htbp]
  \centering
  \includegraphics[scale=0.6]{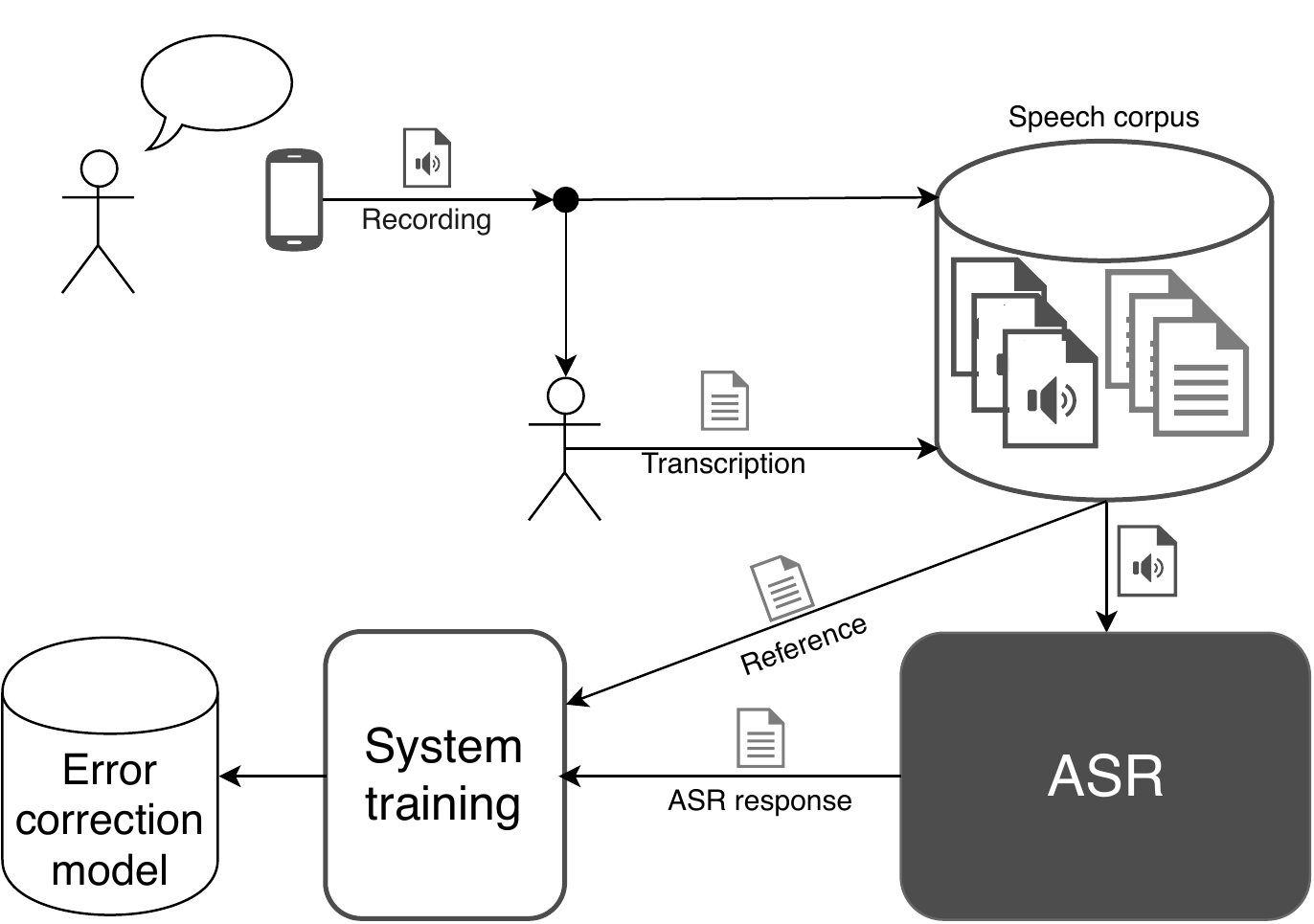}
  \caption{Error correction model training}
\end{figure}

\begin{figure}[htbp]
  \centering
  \includegraphics[scale=0.6]{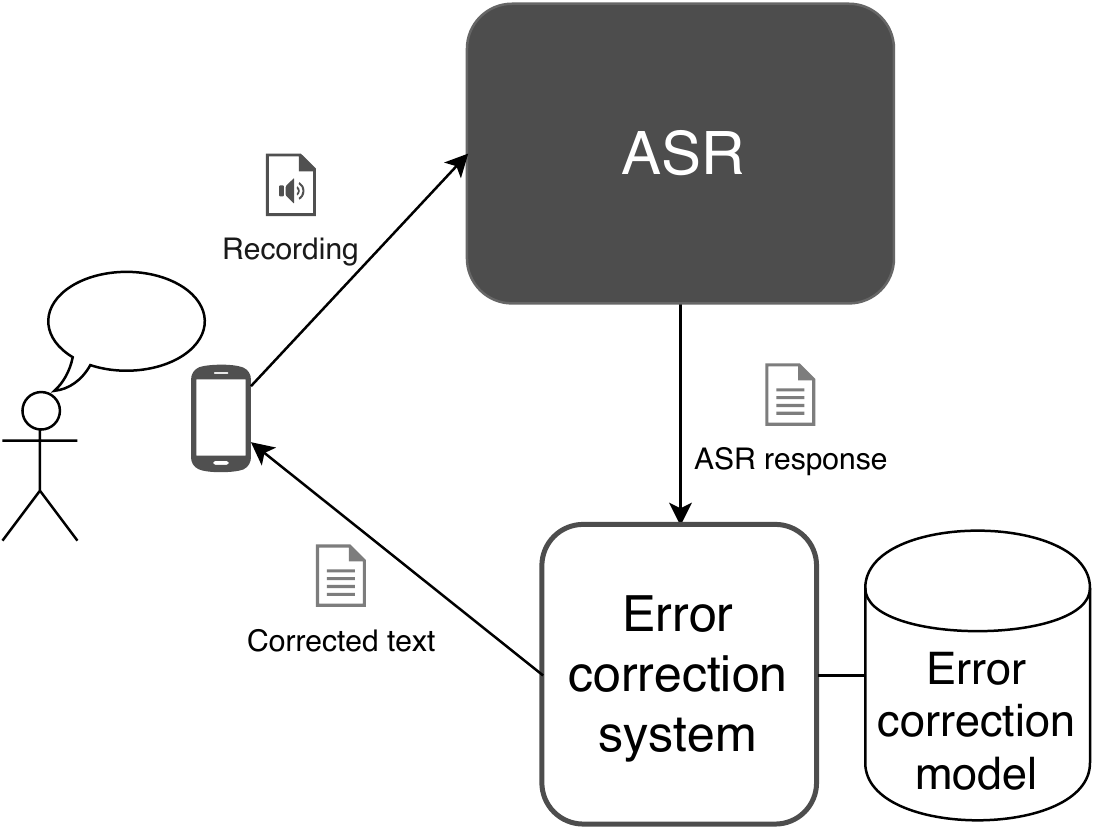}
  \caption{ASR error correction}
\end{figure}

\section{Related work}
\subsection{Shared tasks}
To our best knowledge this is the first open ASR error correction task. However, there were plenty of tasks targeting similar problems. They can be divided into two categories: speech translation tasks and grammatical error correction tasks.
Representatives of the former category are 2 of 3 tasks conducted at 7th International Workshop on
Spoken Language Translation  \cite{iwslt10:EC:overview}. Tasks 1 and 2 provided sentences in source
language in two forms: ASR recognition results (with errors) and correct recognition results
(transcription without errors). Task of the participants was to translate the source text in both
forms to the target language. Participants were provided with 3 training corpora composed of 86225 (Task 1), 19972 and 10061 (Task 2) sentence pairs.
\textit{CoNLL-2013 Shared Task on Grammatical Error Correction} \cite{ng_conll-2013_2013} and
\textit{BEA 2019 Shared Task: Grammatical Error Correction} \cite{BEA2019} are examples of the
second group of tasks. In these tasks participants are given parallel corpora of texts written by
native or non-native English students, containing grammatical, punctuation or spelling errors and
their manually corrected versions. The goal of the proposed system is to correct previously unseen texts. Training corpora in these tasks consist of 38785 and 57151 pairs of sentences respectively.
\subsection{ASR error correction systems}
\cite{ASREDC:review} provide review of some ASR error detection and correction systems together with description of ASR evaluation metrics.
\cite{cucu_statistical_2013} propose error correction using SMT (Statistical Machine Translation)
model. The SMT model is trained on relatively small parallel corpus of 2000 ASR transcripts and
their manually corrected versions. At evaluation time the model is used to ``translate'' ASR
hypothesis into it's corrected form. The system achieves $ 10.5 \% $ relative WER\footnote{Word
Error Rate, see Section \ref{rel_wer}} improvement  by reducing the baseline ASR system's 
WER from $ 11.4 $ to $ 10.2 $ .
\cite{guo2019spelling} describe ASR error correction model based on LSTM sequence-to-sequence
neural network trained on large (40M utterances) speech corpus generated from plain-text data with
text to speech (TTS) system. In addition to the spelling correction model authors experiment with
improving results of end-to-end ASR system by incorporating the external language model and with combination of the two approaches. The proposed system achieves good results ($19\%$ relative WER improvement and $29\%$ relative WER improvement with additional LM re-scoring, with baseline ASR WER of $6.03$ ) but requires large speech corpus or high-quality TTS system to generate such corpus from plain text.

\section{Dataset description}
In order to develop the dataset for the task
we decided to select 9142 sentences from Polish Wikinews \cite{Wikinews19} and 
ask two native speakers of Polish (male and female) to read them
to the speech recognition system. 
Dataset samples consist of transcription of the sentence
being read juxtaposed with 
the textual output captured from the system.
Both references and hypotheses are normalized according to the following rules:
\begin{itemize}
	\item words are uppercased,
	\item all punctuation marks except hyphens are removed,
	\item numbers and special characters are replaced by their spoken forms.
\end{itemize}

The dataset is divided into two sets. 
The training set consists of 8142 utterances randomly sampled from the dataset.
The test set contains the rest of the samples.

The training and test set items consist of:
\begin{enumerate}
  \item \verb|id|: sample identifier
  \item \verb|hyp|: ASR hypothesis - recognition result for the sample voice recording
  \item \verb|ref|: reference - human transcription of the sample recording,
  \item \verb|source|: copyright, source and author attribution information.
\end{enumerate}
Exemplary dataset items are shown in the appendix.
The entire training set and test set (with the exception of reference utterances) are available for
download via Gonito online competition platform \cite{gonito}.

\begin{table}
\begin{tabular}{|l|c|c|}
\hline 
  & Train set & Test set \\ 
\hline 
Number of sentences & 8142 & 1000 \\ 
\hline 
Avarage WER & 3.94 & 4.01 \\ 
\hline 
Sentence Error Rate & 0.74 & 0.75 \\ 
\hline 
Avarage utterance length (words) & 15.40 & 15.10  \\ 
\hline 
Minimum utterance length (words) & 2 & 3 \\ 
\hline 
Maximum utterance length (words) & 100 & 48  \\ 
\hline 
\end{tabular}
\caption{Datasets statistics.}
\end{table}

\begin{figure}[htbp]
  \centering
    \includegraphics[scale=0.4]{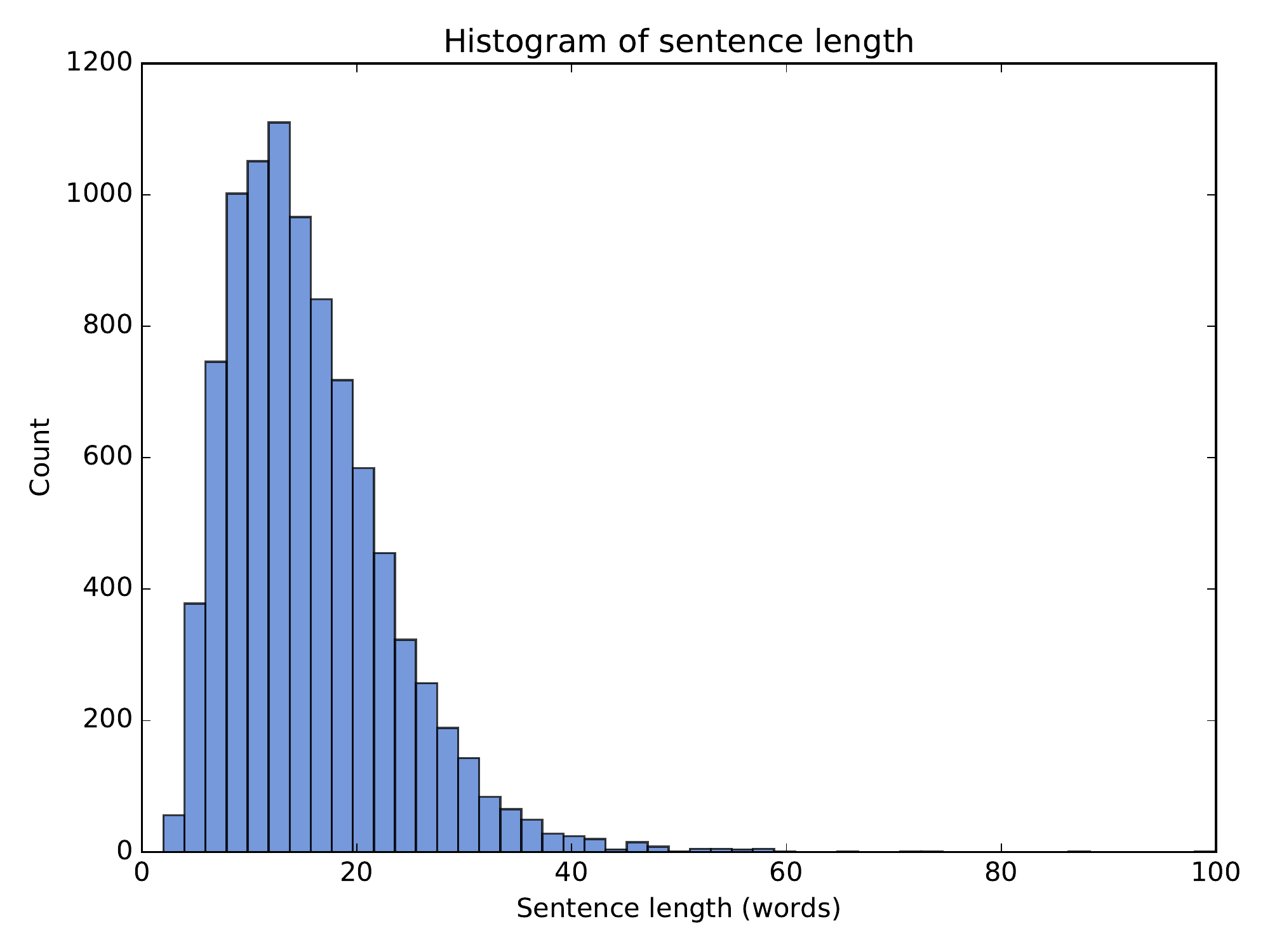}
  \caption{Histogram of sentence length}
\end{figure}

\begin{figure}[htbp]
  \centering
    \includegraphics[scale=0.4]{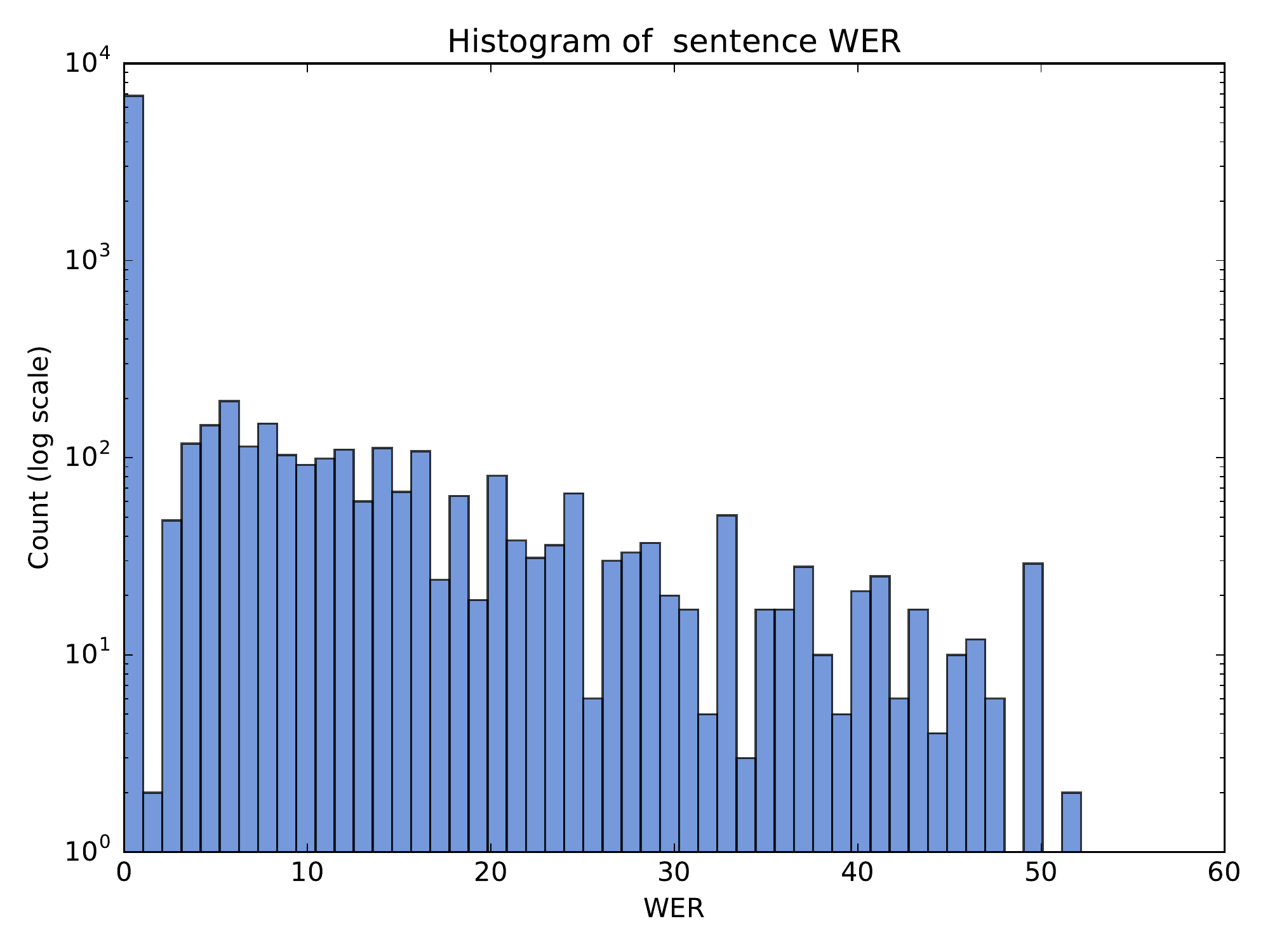}
  \caption{Histogram of Word Error Rate}
\end{figure}

\section{Evaluation}
For the purpose of evaluation the contestants are asked to submit 
their results using the Gonito \cite{gonito} challenge available at \url{https://gonito.net/challenge/asr-corrections}.

The submission consists of single \textit{out.tsv} file containing result of running the proposed system on \textit{in.tsv} file,
containing ASR system output. Both files contain one sentence per line. The output file should be aligned with the input.

The submissions are evaluated using geval tool \cite{geval}, part of the Gonito platform available also as a standalone tool.
Submissions are evaluated using three metrics:
\begin{itemize}
\item \label{rel_wer} WER - Word Error Rate of hypothesis corrected by the proposed system, averaged over all tests sentences.
WER is defined as follows:
\[WER = \frac{S + D + I}{N = H + S + D}\]
where: $S$ = number of substitutions, $D$ = number of deletions, $I$ = number of insertions, $H$ - number of hits, $N$ - length of reference sentence. See \cite{Morris2004FromWA} for in-depth explanation.
\item SRR - Sentence Recognition Rate - sentence level accuracy of hypothesis corrected by the proposed system.
SRR is defined as ratio of the number of sentences with $WER = 0.0$ (correctly recognized
sentences) to the number of all sentences in the corpus.
\item CharMatch - $ F_{0.5} $ - introduced in \cite{jassem-17}. $F_{0.5}$-measure defined in as follows:
  $$ F_{0.5} = (1+0.5^2) \times \frac{P \times R}{0.5^2 P + R} $$
Where: $ P $ is precision and $ R $ is recall:
$$ P = \frac{\sum_i T_i}{\sum_i d_L(h_i,s_i)} , R = \frac{\sum_i T_i}{\sum_i d_L(h_i,r_i)} $$
Where: $r_i$ - i-th reference utterance, $h_i$ - i-th ASR hypothesis,  $s_i$ - i-th system output, $d_L(a,b)$ - Levenshtein distance between sequences $a$ and $b$, $ T_i $ - number of correct changes performed by the system, calculated as:
$$ T_i = \frac{d_L(h_i,r_i) + d_L(h_i, s_i) - d_L(s_i, r_i)}{2} $$

\end{itemize}

\bibliographystyle{ltc05}
\bibliography{ltc-task} 

\onecolumn

\section*{Appendix}
\label{appendix}
	
\begin{table}[ht]
\begin{center}
	\begin{tabular}{|r|p{15cm}|}
  \hline

	\input{example.horizontal.tex}

	\hline
	\end{tabular}
\end{center}

\caption{Dataset samples}
\end{table}

\end{document}

%% file: example.horizontal.tex
\textbf{id}&train-1 \\
\textbf{hypothesis}&DWUDZIESTEGO CZWARTEGO KWIETNIA BIEŻĄCEGO ROKU ROZMAWIALI O WIKIPEDII INTERNECIE WSPÓŁPRACY KLASYFIKOWANIU WIEDZY KSIĄŻKACH I WŁASNOŚCI \textbf{LEKTURA LNEJ} \\
 \textbf{reference}&DWUDZIESTEGO CZWARTEGO KWIETNIA BIEŻĄCEGO ROKU ROZMAWIALI O WIKIPEDII INTERNECIE WSPÓŁPRACY KLASYFIKOWANIU WIEDZY KSIĄŻKACH I WŁASNOŚCI \textbf{INTELEKTUALNEJ} \\
\textbf{source}&https://pl.wikinews.org/w/index.php?curid=27343\&actionaction=history \\
\hline
\textbf{id}&train-2 \\
\textbf{hypothesis}&\textbf{EUROPA POWINNA JĄ} TEŻ ŻE SESJE PE W STRASBURGU SĄ DLA NICH UTRUDNIENIEM BO KOMISJA EUROPEJSKA I RADA UE Z KTÓRYMI PE CIĄGLE WSPÓŁPRACUJE MAJĄ SWOJE STAŁE SIEDZIBY W BRUKSELI \\
\textbf{reference}&\textbf{EUROPOSŁOWIE PRZYPOMINAJĄ} TEŻ ŻE SESJE PE W STRASBURGU SĄ DLA NICH UTRUDNIENIEM BO KOMISJA EUROPEJSKA I RADA UE Z KTÓRYMI PE CIĄGLE WSPÓŁPRACUJE MAJĄ SWOJE STAŁE SIEDZIBY W BRUKSELI \\
\textbf{source}&https://pl.wikinews.org/w/index.php?curid=21290\&actionaction=history \\
\hline
\textbf{id}&train-3 \\
\textbf{hypothesis}&DZIESIĄTEGO WRZEŚNIA DWA TYSIĄCE ÓSMEGO ROKU \textbf{LECH MAM BLADES} OGŁOSIŁ WYNIKI FINANSOWE TRZECIEGO KWARTAŁU WYNOSZĄCE TRZY I \textbf{DZIEWIĘĆDZIESIĄTYCH} MILIARDA DOLARÓW STRAT \\
 \textbf{reference}&DZIESIĄTEGO WRZEŚNIA DWA TYSIĄCE ÓSMEGO ROKU \textbf{LEHMAN BROTHERS} OGŁOSIŁ WYNIKI FINANSOWE TRZECIEGO KWARTAŁU WYNOSZĄCE TRZY I \textbf{DZIEWIĘĆ DZIESIĄTYCH} MILIARDA DOLARÓW STRAT \\
\textbf{source}&https://pl.wikinews.org/w/index.php?curid=25282\&actionaction=history \\
\hline
\textbf{id}&train-4 \\
\textbf{hypothesis}&POCHÓD ROZPOCZĄŁ SIĘ NA PLACU SENATORSKIM ALKAMISTA SENAATINTORILLA A \textbf{PIERWSZE} W SZEREGU SZŁA SZKOŁA TAŃCA SAMBY SAMBIC TANSSIKOULU \\
 \textbf{reference}&POCHÓD ROZPOCZĄŁ SIĘ NA PLACU SENATORSKIM ALKAMISTA SENAATINTORILLA A \textbf{PIERWSZA} W SZEREGU SZŁA SZKOŁA TAŃCA SAMBY SAMBIC TANSSIKOULU \\
\textbf{source}&https://pl.wikinews.org/w/index.php?curid=30303\&actionaction=history \\
\hline
\textbf{id}&train-5 \\
\textbf{hypothesis}&DZIESIĄTEGO PAŹDZIERNIKA W KATOWICKIM SPODKU ODBĘDZIE SIĘ DWUDZIESTA DZIEWIĄTA EDYCJA RAWA BLUES FESTIVAL NAJWIĘKSZEJ BLUESOWEJ IMPREZY TYPU INDOOR W EUROPIE \\
 \textbf{reference}&DZIESIĄTEGO PAŹDZIERNIKA W KATOWICKIM SPODKU ODBĘDZIE SIĘ DWUDZIESTA DZIEWIĄTA EDYCJA RAWA BLUES FESTIVAL NAJWIĘKSZEJ BLUESOWEJ IMPREZY TYPU INDOOR W EUROPIE \\
\textbf{source}&https://pl.wikinews.org/w/index.php?curid=25476\&actionaction=history \\
\hline

\textbf{id}&train-6 \\
\textbf{hypothesis}&PRZEPROWADZONE W POŁOWIE GRUDNIA DWUSTRONNE ROZMOWY NIE PRZYNIOSŁY REZULTATU \\
\textbf{reference}&PRZEPROWADZONE W POŁOWIE GRUDNIA DWUSTRONNE ROZMOWY NIE PRZYNIOSŁY REZULTATU \\
\textbf{source}&https://pl.wikinews.org/w/index.php?curid=5050\&actionaction=history \\
\hline